\def\year{2021}\relax
\documentclass[letterpaper]{article} 
\usepackage{aaai21}  
\usepackage{times}  
\usepackage{helvet} 
\usepackage{courier}  
\usepackage[hyphens]{url}  
\usepackage{graphicx} 
\usepackage{natbib}  
\usepackage{caption} 
\usepackage{array}
\newcolumntype{P}[1]{>{\centering\arraybackslash}p{#1}}
\newcommand\norm[1]{\left\lVert#1\right\rVert}
\usepackage{graphicx}
\usepackage{subcaption}
\usepackage[ruled]{algorithm2e}
\usepackage{amsmath}
\usepackage{amsfonts}
\usepackage{setspace}
\usepackage{booktabs}

\title{Interpretable Graph Capsule Networks for Object Recognition}

\title{Interpretable Graph Capsule Networks for Object Recognition}
\author {
        Jindong Gu, Volker Tresp \\
}

\affiliations {
    University of Munich  \\
    Siemens AG, Corporate Technology  \\
    jindong.gu@outlook.com
}

\begin{document}

\maketitle

\begin{abstract}
Capsule Networks, as alternatives to Convolutional Neural Networks, have been proposed to recognize objects from images. The current literature demonstrates many advantages of CapsNets over CNNs. However, how to create explanations for individual classifications of CapsNets has not been well explored. The widely used saliency methods are mainly proposed for explaining CNN-based classifications; they create saliency map explanations by combining activation values and the corresponding gradients, e.g., Grad-CAM. These saliency methods require a specific architecture of the underlying classifiers and cannot be trivially applied to CapsNets due to the iterative routing mechanism therein. To overcome the lack of interpretability, we can either propose new post-hoc interpretation methods for CapsNets or modifying the model to have build-in explanations. In this work, we explore the latter. Specifically, we propose interpretable Graph Capsule Networks (GraCapsNets), where we replace the routing part with a multi-head attention-based Graph Pooling approach. In the proposed model, individual classification explanations can be created effectively and efficiently. Our model also demonstrates some unexpected benefits, even though it replaces the fundamental part of CapsNets. Our GraCapsNets achieve better classification performance with fewer parameters and better adversarial robustness, when compared to CapsNets. Besides, GraCapsNets still keep other advantages of CapsNets, namely, disentangled representations and affine transformation robustness.
\end{abstract}

\section{Introduction}
In past years, Convolutional Neural Networks (CNNs) have become the standard model applied in object recognition. Our community has been pursuing more powerful CNN models with compact size \cite{he2016deep}. Besides, two weaknesses of CNNs have also been intensively investigated recently. Namely, 1) Adversarial Vulnerability \cite{Szegedy2014IntriguingPO}: The predictions of CNNs can be misled by imperceptible perturbations of input images. 2) Lack of Interpretability \cite{Simonyan2013DeepIC}: The predictions of standard CNNs are based on highly entangled representations. The two weaknesses might be attributed to the fact that the representations learned by CNNs are not aligned to human perception.

Recently, Capsule Networks (CapsNets) \cite{sabour2017dynamic} have been proposed and received much attention since they can learn more human-aligned visual representations \cite{qin2019detecting}. The disentangled representations captured by CapsNets often correspond to human-understandable visual properties of input objects, e.g., rotations and translations. Recent work on CapsNets aims to propose more efficient routing algorithms \cite{e2018matrix,NIPS2019_8982,zhang2018cappronet,Tsai2020Capsules} and understand the contributions of the routing algorithms \cite{gu2019improving,gu2021effective}.

However, how to explain individual classifications of CapsNets has been less explored. The state-of-the-art saliency methods are mainly proposed for CNNs, e.g., Grad-CAM \cite{selvaraju2017grad}. They combine activation values and the received gradients in specific layers, e.g., deep convolutional layers. In CapsNets, instead of deep convolutional layers, an iterative routing mechanism is applied to extract high-level visual concepts. Hence, these saliency methods cannot be trivially applied to CapsNets. Besides, the routing mechanism makes it more challenging to identify interpretable input features relevant to a classification.

In this work, we propose interpretable Graph Capsule Networks (GraCapsNets). In CapsNets, the primary capsules represent object parts, e.g., eyes and nose of a cat. In our GraCapsNets, we explicitly model the relationship between the primary capsules (i.e., part-part relationship) with graphs. Then, the followed graph pooling operations pool relevant object parts from the graphs to make a classification vote. Since the graph pooling operation reveals which input features are pooled as relevant ones, we can easily create explanations to explain the classification decisions. Besides the interpretability, another motivation of GraCapsNets is that the explicit part-part relationship is also relevant for object recognition, e.g., spatial relationships.

The classic graph pooling algorithms are clustering-based, which requires high computational complexity. It is challenging to integrate these graph pooling algorithms into neural networks. Recent progress on graph pooling modules of Graph Neural Networks makes similar integrations possible. E.g., \cite{Ying2018HierarchicalGR} proposed a differentiable graph pooling module, which can be integrated into various neural network architectures in an end-to-end fashion.

The capsule idea is also integrated into Graph Neural Networks for better graph classification \cite{verma2018graph,xinyi2018capsule}. They treat node feature vectors as primary capsules and aggregates information from the capsules via a routing mechanism. Different from their works, we integrate graph modeling into CapsNets for better object recognition. On the contrary, our GraCapsNets treat capsules as node feature vectors and represent them as graphs so that we can leverage graph structure information (e.g., the spatial part-part relationship between object parts).

Our main contribution of this work is to propose GraCapsNets, where we replace the fundamental routing part of CapsNets with multi-head attention-based Graph Pooling operations. On GraCapsNets, we can create explanations for individual classifications effectively and efficiently. Besides, our empirical experiments show that GraCapsNets achieve better performance with fewer parameters and also learn disentangled representations. GraCapsNets are also shown to be more robust to the primary white adversarial attacks than CNNs and various CapsNets.

\section{Related Work}
\textbf{Routing Mechanism:} The goal of routing processes in CapsNets is to identify the weights of predictions made by low-level capsules, called coupling coefficients (CCs) in \cite{sabour2017dynamic}. Many routing mechanisms have been proposed to improve Dynamic Routing \cite{sabour2017dynamic}; they differ from each other only in how to identify CCs.

Dynamic Routing \cite{sabour2017dynamic} identifies CCs with an iterative routing-by-agreement mechanism. EM Routing \cite{e2018matrix} updates CCs iteratively using the Expectation-Maximization algorithm. \cite{chen2018generalized} removes the computationally expensive routing iterations by predicting CCs directly. To improve the prediction of CCs further, Self-Routing \cite{NIPS2019_8982} predicts CCs using a subordinate routing network. However, \cite{gu2019improving} shows that similar performance can be achieved by simply averaging predictions of low-level capsules without learning CCs. In this work, we propose Graph Capsule Networks, where a multi-head attention-based graph pooling mechanism is used instead of routing.

\textbf{Graph Pooling:} Earlier works implement graph pooling with clustering-based graph coarsening algorithms, e.g., Graclus \cite{Dhillon2007WeightedGC}, where the nodes with similar representations are clustered into one. In later works (Set2Set \cite{Vinyals2016OrderMS} and SortPool \cite{zhang2018end}), the graph features are also taken into consideration. However, they require the ordering of the nodes by a user-defined meaningful criterium. Recently, the seminal work \cite{Ying2018HierarchicalGR} proposes a differentiable graph pooling module, which can be combined with various neural network architectures in an end-to-end fashion. For simplification of \cite{Ying2018HierarchicalGR}, top-K pooling \cite{graphunet,understand2019gnn} and self-attention pooling \cite{lee2019self} have been proposed. Almost all the graph pooling strategies have been mainly used for graph classification. Based on the work \cite{Ying2018HierarchicalGR}, we propose multiple-heads attention-based graph pooling for object recognition.

\textbf{Adversarial Robustness:} \cite{Szegedy2014IntriguingPO} shows that imperceptible image perturbations can mislead standard CNNs. Since then, many adversarial attack methods have been proposed, e.g., FGSM \cite{Goodfellow2015ExplainingAH}, C\&W \cite{Carlini2017Robust}. Meanwhile, the approaches to defend these attacks have also been widely investigated, e.g., Adversarial Training \cite{Madry2017TowardsDL,Athalye2018ObfuscatedGG}, Certified Defenses \cite{wong2017provable,Cohen2019CertifiedAR}. One way to tackle the adversarial vulnerability is to propose new models that learn more human perception-aligned feature representations, e.g., CapsNets \cite{sabour2017dynamic,qin2019detecting}. Recent work \cite{e2018matrix,NIPS2019_8982} shows that CapsNets with their routing processes are more robust to white-box adversarial attacks.

\textbf{Interpretability:} A large number of interpretation methods have been proposed to understand individual classifications of CNNs. Especially, saliency maps created by post-hoc methods, as intuitive explanations, have received much attention. We categorize the methods into two categories. The first category is architecture-agnostic, such as, vanilla Gradients (Grad) \cite{Simonyan2013DeepIC}, Integrated Gradients (IG) \cite{sundararajan2017axiomatic} as well as their smoothed versions (SG) \cite{smilkov2017smoothgrad}. The second one requires specific layers or architecture of models, e.g., Guided Backpropagation \cite{springenberg2014striving,gu2019saliency}, DeepLIFT \cite{shrikumar2017learning}, LRP \cite{bach2015pixel,gu2018understanding}, Grad-CAM \cite{selvaraju2017grad}. Only the architecture-agnostic methods can be trivially generalized to CapsNets due to the routing mechanism therein. In our GraCapsNets, the explanations can be created with attention in the graph pooling operations.
  
\begin{figure*}[t]
  \centering
   \includegraphics[width=0.85\linewidth]{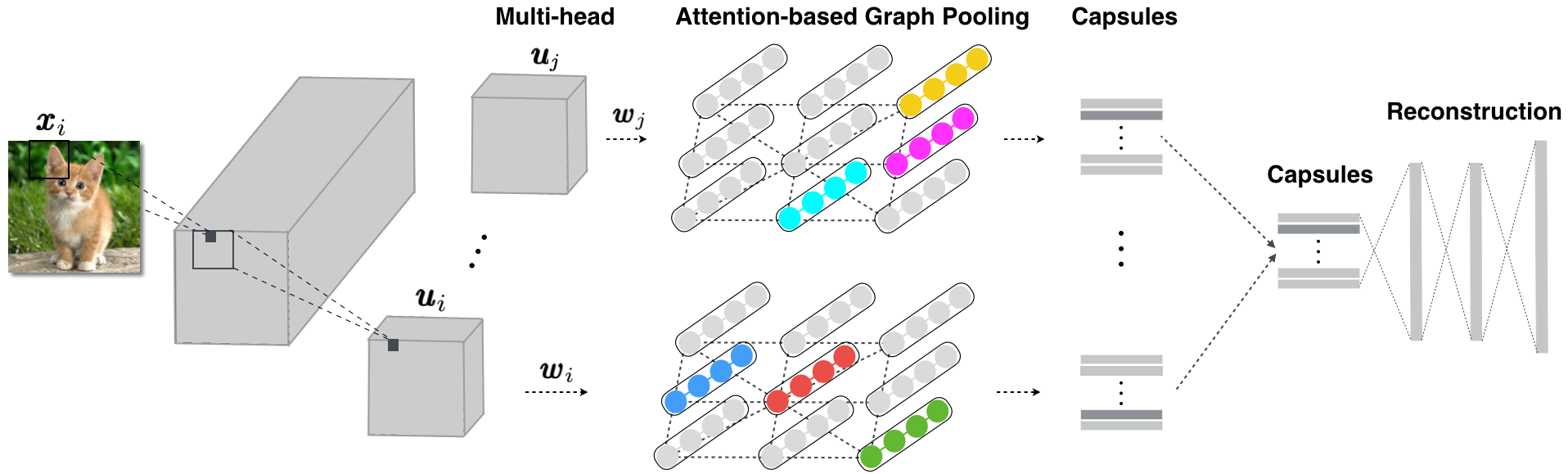}
  \caption{The illustration of GraCapsNets: The extracted primary capsules are transformed and modeled as multiple graphs.  The pooling result on each graph (head) corresponds to one vote. The votes on multiple graphs (heads) are averaged to generate the final prediction.}
  \label{fig:overview}
\end{figure*}

 \begin{algorithm}[t]
\setstretch{1.18}
\small
\SetAlgoLined
\textbf{Input:} An image $\mathbf{X}$  \\
\textbf{Output:} Class capsules $\mathbf{V}\in\mathbb{R}^{M\times D_{out}}$  \\
$1.$ Extract primary capsules $\mathbf{u}_i\in\mathbb{R}^{D_{in}}$ from input $\mathbf{X}$\;
$2.$ Transform each $\mathbf{u}_i$ into $\mathbf{\hat{u}}_{j|i}\in\mathbb{R}^{D_{out}}$\;
$3.$ Identify all $c_{ij}$ with a routing process\;
$4.$ Compute $\mathbf{s}_j = \sum^N_{i=1} c_{ij} * \mathbf{\hat{u}}_{j|i}$\;
$5.$ Output capsules $\mathbf{v}_j = squash(\mathbf{s}_j)$\\
\caption{Capsule Networks}
\label{alg:capsnet}
\end{algorithm}

\section{Graph Capsule Networks}
We first briefly review CapsNets. As shown in Algorithm \ref{alg:capsnet}, CapsNets start with convolutional layers that convert the input pixel intensities $\mathbf{X}$ into primary capsules $\mathbf{u}_i$ (i.e., low-level visual entities). Each $\mathbf{u}_i$ is transformed to vote for high-level capsules $\mathbf{\hat{u}}_{j|i}$ with learned transformation matrices. Then, a routing process is used to identify the coupling coefficients $c_{ij}$, which describe how to weight votes from primary capsules. Finally, a squashing function is applied to the identified high-level capsules $\mathbf{s}_j$ so that the lengths of them correspond to the confidence of the class's existence. A reconstruction part works as regularization during training.
 
Different routing mechanisms differ only in the 3rd step, i.e., how to identify $c_{ij}$. Routing processes describe one way to aggregate information from primary capsules into high-level ones. In our GraCapsNets, we implement the information aggregation by multi-head graph pooling processes.

\begin{algorithm}[t]
\setstretch{1.2}
\small
\SetAlgoLined
\textbf{Input:} An image $\mathbf{X}$ \\
\textbf{Output:} Class capsules $\mathbf{V}\in\mathbb{R}^{M\times {D_{out}}}$  \\
$1.$ Extract primary capsules $\mathbf{u}_i\in\mathbb{R}^{D_{in}}$ from input $\mathbf{X}$\;
$2.$ Project each $\mathbf{u}_i$ into the feature space $\mathbf{u}^\prime_{i}\in\mathbb{R}^{D_{out}}$\;
$3.$ Model all $\mathbf{u}^\prime_{i}$ as multiple graphs\;
$4.$ Compute $\mathbf{s}_j\in\mathbb{R}^{D_{out}}$ with multi-head graph pooling\;
$5.$ Output capsules $\mathbf{v}_j = squash(\mathbf{s}_j)$ \\
\caption{Graph Capsule Networks}
\label{alg:gracapsnet}
\end{algorithm}

As shown in Algorithm \ref{alg:gracapsnet}, GraCapsNets differ from CapsNets in the steps of 2, 3, and 4. In GraCapsNet, the primary capsules $\mathbf{u}_i$ are transformed into a feature space. All transformed capsules $\mathbf{u}^\prime_{i}$ are modeled as multiple graphs. Each graph corresponds to one head, the pooling result on which corresponds to one vote. The votes on multiple heads are averaged as the final prediction. The GraCapsNets is also illustrated in Figure \ref{fig:overview}. 

In CapsNets, most of the parameters are from the transformation matrix $\mathbf{W}^t\in\mathbb{R}^{N \times D_{in} \times (M\ast D_{out})}$ where $D_{in}, D_{out}$ are the dimensions of input primary capsules and output high-level capsules, $N$ is the number of primary capsules, and $M$ is the number of output classes. In GraCapsNets, the transformation matrix is $\mathbf{W}^t \in\mathbb{R}^{N \times D_{in} \times D_{out}}$ and the trainable parameters in the graph pooling layer is $\mathbf{W} \in \mathbb{R}^{D_{out} \times M}$. Hence, the parameters are reduced signigicantly.

\subsection{Multiple Heads in GraCapsNets}
We now introduce how to model all transformed capsules $\mathbf{u}^\prime_{i}$ as multiple graphs. A graph consists of a set of nodes and a set of edges.

As shown in GraCapsNet in Figure \ref{fig:overview}, the primary capsules are reshaped from $L$ groups of feature maps. Each group consists of $C$ feature maps of the size $K\times K$. Correspondingly, the transformed capsules $\mathbf{u}^\prime_{i}$ where $i \in \{ 1, 2, ... K^2\}$ form a single graph with $K^2$ nodes. Namely, the capsules of the same type (the ones on the same feature maps but different locations) are modeled in the same graph. Each node corresponds to one transformed capsule $\mathbf{u}^\prime_{i}$, and the activation vector of $\mathbf{u}^\prime_{i}$ is taken as features of the corresponding node.

The graph edge information can be represented by an adjacency matrix, in which different priors can be modeled, e.g., camera geometry \cite{Khasanova2019GraphbasedIR} and spatial relationships \cite{knyazev2019image}. In this work, we model the spatial relationship between primary capsules since they can be computed without supervision.

For the above graph with $K^2$ nodes, elements in the adjacency matrix $\mathbf{A} \in\mathbb{R}^{K^2 \times K^2}$ can be computed as
\begin{equation}
A_{ij} =e^{(-\frac{\norm{\mathbf{p}_i - \mathbf{p}_j}^2}{2\sigma ^2} )}
\end{equation}
where $i, j$ are indice of nodes and $\mathbf{p}_i \in\mathbb{R}^{2}, \mathbf{p}_ j \in\mathbb{R}^{2}$ are coordinates of the nodes, i.e. from $(1,1)$ to $(K, K)$. Similarly, we can build $l$ graphs (heads) in total with the same adjcency matrix. They differ from each other in node features.

\begin{table*}[h]
\small
\begin{center}
\begin{tabular}{c|P{1.2cm}|P{1.6cm}|P{1.2cm}|P{1.6cm}|P{1.2cm}|P{1.6cm}}
\hline
Datasets & \multicolumn{2}{c|}{MNIST}  &  \multicolumn{2}{c|}{Fashion MNIST}  &  \multicolumn{2}{c}{CIFAR10}   \\
\hline
Model & \#Para.(M) & Accuracy & \#Para.(M) & Accuracy & \#Para.(M) & Accuracy  \\
\hline
CapsNets \cite{sabour2017dynamic} & 6.54  &99.41{\scriptsize ($\pm$ 0.08)} & 6.54  &  92.12{\scriptsize ($\pm$ 0.29)}  &  7.66  &74.64{\scriptsize ($\pm$ 1.02)}   \\
\hline
\textbf{GraCapsNets} &  \textbf{1.18} & \textbf{99.50}{\scriptsize ($\pm$ 0.09)} & \textbf{1.18} &  \textbf{93.1}{\scriptsize ($\pm$ 0.09)}   & \textbf{2.90} & \textbf{82.21}{\scriptsize ($\pm$ 0.11)}  \\
\hline
\end{tabular}
\end{center}
\caption{Compared to CapsNets, GraCapsNets achieve slightly better performance on grayscale image datasets and significantly better performance on CIFAR10 with fewer parameters.}
\label{tab:perf}
\end{table*}

\subsection{Graph Pooling in GraCapsNets}
\label{sec:pool}
Given node features $\mathbf{X}^l \in \mathbb{R}^{(K^2 \times D_{out})}$ and adjacency matrix $\mathbf{A}\in \mathbb{R}^{(K^2 \times K^2)}$ in the $l$-th head of GraCapsNet, we now describe how to make a vote for the final prediction by a attention-based graph pooling operation. We first compute the attention of the head as
\begin{equation}
\mathbf{Att}^l = \mathrm{softmax}(\mathbf{A} \mathbf{X}^l \mathbf{W})
\end{equation}
where $\mathbf{W} \in\mathbb{R}^{D_{out} \times M} $ are learnable parameters. $D_{out}$ is the dimension of the node features and $M$ is the number of output classes. The output is of the shape $(K^2 \times M)$. In our GraCapsNet for object recognition, $\mathbf{Att}^l$ corresponds to the visual attention of the heads.

The visual attention describes how important each low-level visual entity is to an output class. We normalize attention output with softmax function in the first dimension, i.e., between low-level entities. Hence, the attention on a visual entity could be nearly zero for all classes. Namely, a visual entity can abstain from voting. When some visual entities correspond to the noisy background of the input image, the noise can be filtered out by the corresponding abstentions.

The attention is used to pool nodes of the graph for output classes. The graph pooling output $\mathbf{S}^l \in \mathbb{R}^{(M \times D_{out})}$ of the head is computed as
\begin{equation}
\mathbf{S}^l = (\mathbf{Att}^l)^T \mathbf{X}^l.
\end{equation}

The final predictions of GraCapsNets are based on all $L$ heads with outputs $\mathbf{S}^l$ where $l \in \{1, 2, ..., L\}$. The output capsules are
\begin{equation}
\mathbf{V} = \mathrm{squash}(\frac{1}{L} \sum^L_{l=1} \mathbf{S}^l)
\end{equation}

Following CapsNets \cite{sabour2017dynamic}, the squashing function is applied to each high-level capsule $\mathbf{s}_j \in \mathbb{R}^{D_{out}}$.
\begin{equation}
\mathrm{squash}(\mathbf{s}_j) = \frac{\norm{\mathbf{s}_j}^2}{1+\norm{\mathbf{s}_j}^2} \frac{\mathbf{s}_j}{\norm{\mathbf{s}_j}}
\end{equation}
 and the loss function used to train our GraCapsNets is
\begin{equation}
\begin{split}
L_k = & T_k  \max(0, m^+ - \norm{\mathbf{v}_k})^2 \\
& + \lambda (1 -T_k) \max(0, \norm{\mathbf{v}_k} -  m^-)^2
\end{split}
\end{equation}
where $T_k = 1$ if the object of the $k$-th class is present. As in \cite{sabour2017dynamic}, the hyper-parameters are often empirically set as $m^+ = 0.9$, $m^- = 0.1$ and $\lambda=0.5$.
The effectiveness of Graph Pooling as well as Multiple Heads is verified in the experimental section.

\subsection{Interpretability in GraCapsNets}
\label{sec:exp}
There is no interpretation method designed specifically for CapsNets. The existing ones were proposed for CNNs. Only the architecture-agnostic ones \cite{Simonyan2013DeepIC,sundararajan2017axiomatic,smilkov2017smoothgrad} can be trivially generalized to CapsNets, which only requires the gradients of the output with respect to the input.

In our GraCapsNet, we can use visual attention as built-in explanation to explain the predictions of GraCapsNets. The averaged attenion over $l$ heads is
\begin{equation}
\mathbf{E} = \frac{1}{L} \sum^L_{l=1} \mathbf{Att}^l
\end{equation}
where $\mathbf{Att}^l$ corresponds to the attention of the $l$-th head. The created explanations $\mathbf{E}$ are of the shape $(K^2 \times M)$. Given the predicted class, the $K\times K$ attention map indicates which pixels of the input image support the prediction.

\section{Experiments}
Many new versions of CapsNets have been proposed, and they report competitive classification performance. However, the advantages of CapsNets over CNNs are not only in performance but also in other properties, e.g., disentangled representations, adversarial robustness. Additionally, instead of pure convolutional layers, ResNet backbones\cite{he2016deep} are often applied to extract primary capsules to achieve better performance.

Hence, in this work, we \textbf{comprehensively} evaluate our GraCapsNets from the four following aspects. All scores reported in this paper are averaged over 5 runs.
\begin{enumerate}
\item Classification Performance: Comparison of our GraCapsNets with original CapsNets built on two convolutional layers and the ones built on ResNet backbones.
\item Classification Interpretability: Comparison of explanations in Section \ref{sec:exp} with the ones created by the architecture-agnostic saliency methods.
\item Adversarial Robustness: Comparison of GraCapsNets with various CapsNets and counter-part CNNs.
\item We show GraCapsNets also learn disentangled representations and achieve similar transformation robustness.
\end{enumerate}

\begin{table*}[t]
\begin{center}
\small
\begin{tabular}{ccccP{1.6cm}}
\toprule
Models &  \#Para.(M) & FLOPs(M) & CIFAR10 & SVHN  \\
\midrule
Backbone + Avg  & 0.27 & 41.3  & 7.94($\pm$0.21) &  3.55($\pm$0.11)  \\
Backbone + FC   & 0.89 & 61.0 & 10.01($\pm$0.99) &  3.98($\pm$0.15)  \\
\midrule
Dynamic Routing \cite{sabour2017dynamic} & 5.81 & 73.5 & 8.46($\pm$0.27) &  3.49($\pm$0.69)\\
EM Routing \cite{e2018matrix} & 0.91 & 76.6 & 10.25($\pm$0.45) & 3.85($\pm$0.13)  \\
Self-Routing \cite{NIPS2019_8982} & 0.93 & 62.2 & 8.17($\pm$0.18) & 3.34($\pm$0.08)  \\
\textbf{GraCapsNets} &\textbf{0.28} & \textbf{59.6} & \textbf{7.99}($\pm$0.13) & \textbf{2.98}($\pm$0.09)  \\
\bottomrule
\end{tabular}
\end{center}
 \caption{Comparison to state-of-the-art CapsNets performance on the benchmark datasets.}
\label{tab:SOTA_perf}
\end{table*}

\subsection{Classification Performance}
\label{sec:basic}
The datasets, MNIST \cite{lecun1998gradient}, F-MNIST \cite{xiao2017online} and CIFAR10 \cite{krizhevsky2009learning}, are used in this experiment. The data preprocessing, the arhictectures and the training procedure are set identically to \cite{sabour2017dynamic} (See Supplement A). Correspondingly, in GraCapsNets, $32$ heads and $8D$ primary capsules are used. $3\times 3$ kernels are used in Conv layers to obtain graphs with 144 nodes on MNIST, 196 nodes on CIFAR10.

\textbf{Comparison with the original CapsNets} The classification results are reported in Table \ref{tab:perf}. In grayscale images, GraCapsNets achieve slightly better performance with fewer parameters. In CIFAR10, our model outperforms the CapsNet by a large margin. The reason behind this is that our graph pooling process can better filter out the background noise. The pixel values of the background of grayscale images are often zeros, not noisy. Hence, our model performs much better on realistic datasets.

\begin{figure}[h]
    \centering
    \begin{subfigure}[b]{0.5\textwidth}
     \centering
      \hspace{-0.7cm}
        \includegraphics[scale=0.38]{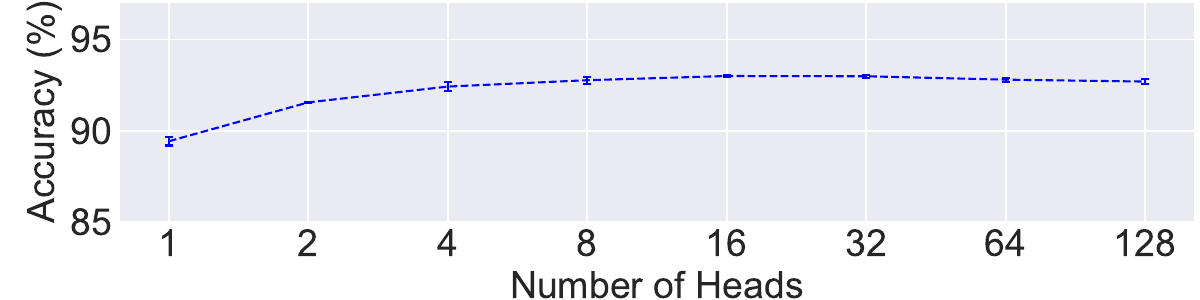}
        \caption{F-MNIST Dataset}
    \end{subfigure}
    \begin{subfigure}[b]{0.5\textwidth}
     \centering
     \hspace{-0.7cm}
        \includegraphics[scale=0.38]{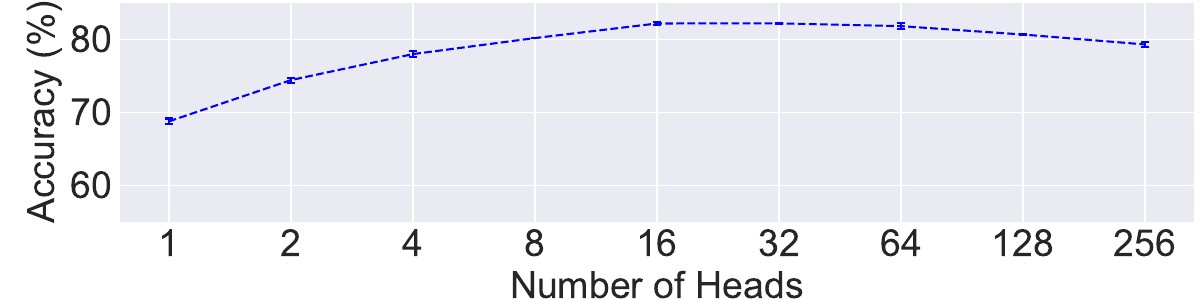}
        \caption{CIFAR10 Dataset}
    \end{subfigure}
    \caption{Ablation study on multiple heads: Given fixed channels, the GraCapsNets with more heads perform better in general. The GraCapsNets with too many heads can degrade a little since the small primary capsules are not able to represent visual entities well.}
    \label{fig:ab_heads}
\end{figure}

\textbf{Ablation Study on Multiple Heads} In this experiment, we set the number of feature maps fixed (e.g., 256 on F-MNIST). We train GraCapsNets with different number of heads $2^n$ where $n \in \{0, 1, ... 7\}$. The corresponding dimensions of the primary capsules are $2^n$ where $n \in \{8, 7, ... 1\}$. The performance is shown in Figure \ref{fig:ab_heads}. The GraCapsNet with more heads achieves better performance in general. However, when too many heads are used, the performance decreases a little. In that case, the dimensions of the primary capsules are too small to capture the properties of low-level visual entities. Overall, our model is not very sensitive to the number of heads. When the number heads vary from $16$ to $64$, our models show similar performance with tiny variance.

\begin{figure}[h]
    \centering
    \begin{subfigure}[b]{0.5\textwidth}
    \centering
    \hspace{-0.6cm}
        \includegraphics[scale=0.39]{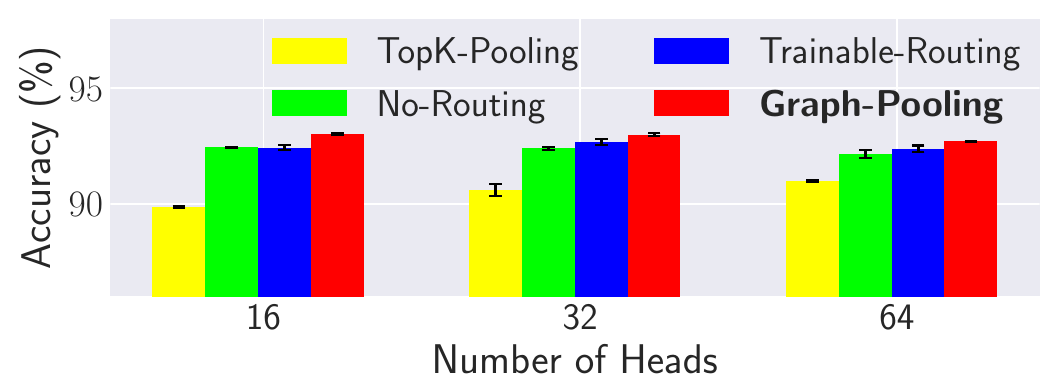}
        \caption{F-MNIST Dataset}
    \end{subfigure}
    \begin{subfigure}[b]{0.5\textwidth}
    \centering
    \hspace{-0.6cm}
        \includegraphics[scale=0.39]{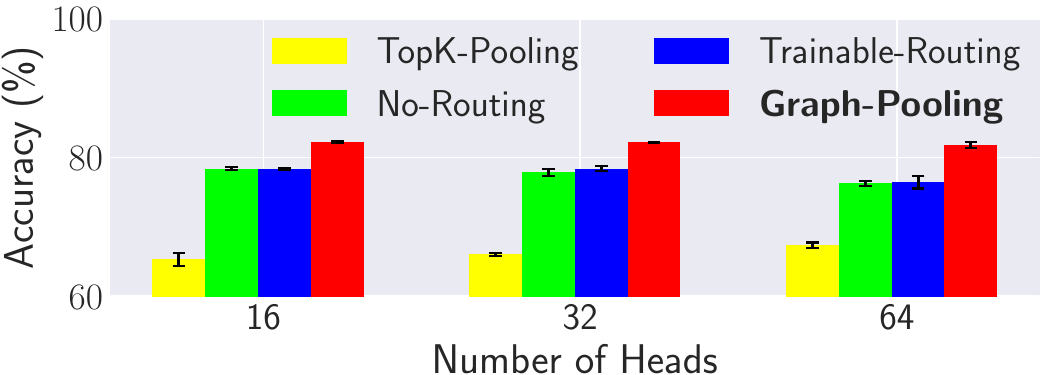}
        \caption{CIFAR10 Dataset}
    \end{subfigure}
    \caption{Ablation Study on Graph Pooling: GraCapsNets with graph modeling outperform others.}
    \label{fig:ab_pooling}
\end{figure}

\textbf{Ablation Study on Graph Pooling} In GraCapsNets, we model the transformed capsules as multiple graphs. The spatial relationship between the transformed capsules is modeled in each graph. To investigate the effectiveness of the graph modeling, we compare GraCapsNets with closely related pooling operations as well as routing mechanisms.

Top-K graph pooling \cite{graphunet,understand2019gnn}, simplified version of our graph pooling approach, projects node features into a feature space, and chooses the top-K ones to coarsen the graph, where the graph structure (spatial relationship) is not used. In addition, the trainable routing algorithm \cite{chen2018generalized} predict directly which primary capsules should be routed to which output capsules. In No-routing algorithm \cite{gu2019improving}, the transformed capsules are simply averaged to obtain output capsules. The two routing algorithms are strong baselines and leverage no graph information when aggregating information.

We report the performance of different graph pooling operations and routing algorithms in Figure \ref{fig:ab_pooling}. Our Graph-Pooling with different heads outperforms others on both datasets, which indicate the effectiveness of the part-part relationship modeled in our Graph-Pooling process.

\begin{figure*}[t]
    \centering
    \begin{subfigure}[b]{\textwidth}
     \centering
        \includegraphics[width=0.98\textwidth]{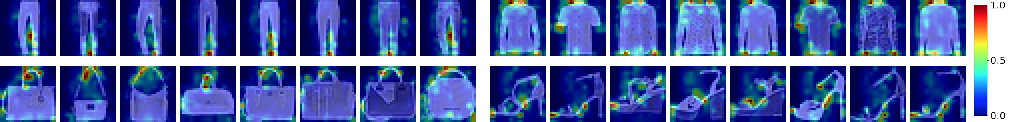}
        \caption{Visual Attention as Explanations on F-MNIST Dataset.}
        \label{fig:fmnist_exp}
    \end{subfigure}
    \begin{subfigure}[b]{\textwidth}
     \centering
        \includegraphics[width=0.98\textwidth]{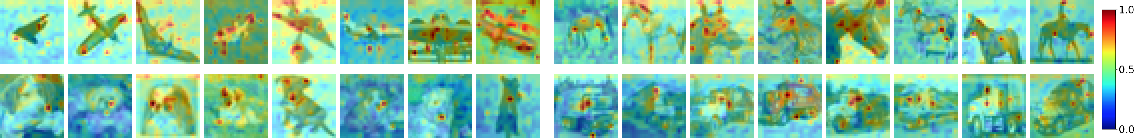}
        \caption{Visual Attention as Explanations on CIFAR10 Dataset.}
        \label{fig:cifar10_exp}
    \end{subfigure}
    \caption{Visual Attention in GraCapsNets: the models focus on discriminative input visual features, e.g., the handles of the handbags and the wings of the planes.}
    \label{fig:quali_exps}
\end{figure*}

\textbf{Comparison with various CapsNets on ResNet Backbones} \label{sec:sota}
The backbones are supposed to extract more accurate primary capsules. To compare with various CapsNets, we also build our GraCapsNets on their backbones. Following \cite{NIPS2019_8982}, we apply Dynamic routing, EM-routing, Self-routing, and our Multi-head Graph Pooling on the ResNet20 \cite{he2016deep} backbone. Two CNN baselines are Avg): the original ResNet20 and FC): directly followed by Conv + FC without pooling.

The performance is reported in Table \ref{tab:SOTA_perf}. Our GraCapsNets outperform previous routing algorithm slightly, but with fewer parameters and less computational cost. Our GraCapsNets achieve better performance than similar-sized CNNs. The size of GraCapsNets is even comparable to the original ResNet20.  Besides the popular routing mechanisms above, other new CapsNets architectures \cite{ahmed2019star} and Routing mechanisms \cite{zhang2018cappronet,Tsai2020Capsules} have also been recently proposed. They report scores on different backbones in different settings. Compared to scores reported in their papers, ours also achieves comparable performance with fewer parameters.

\subsection{Classification Interpretability}
The predictions of GraCapsNet can be easily explained with their visual attention. We visualize the attention in inferences and compare them with the explanations created by other appliable interpretation methods, namely, Grad \cite{Simonyan2013DeepIC}, IG \cite{sundararajan2017axiomatic}, Grad-SG and IG-SG \cite{smilkov2017smoothgrad}. In this experiment, the settings of these methods follow Captum package \cite{captum2019github} (See Supplement B). Only GraCapsNets are used. We use the ones with basic architecture from Section \ref{sec:basic}.

\textbf{Qualitative Evaluation} We make predictions with our GraCapsNets for some examples chosen randomly from test datasets. The visual attention is visualized on the original input in Figure \ref{fig:quali_exps}. The color bars right indicate the importance of the input features, where blue corresponds to little relevance, dark red to high relevance.

For instance, in F-MNIST, the trouser legs and the gap between them are relevant for the recognition of the class \textit{Trouser}, the handles is to \textit{Bag}; In CIFAR10, the wings to \textit{Plane}, and the heads (especially the noses) to \textit{Dog}. Since the visual attention is more aligned with human-vision perception, the observations also explain why our models are more robust to adversarial examples. We also visualize explanations created by all baseline methods, which are less interpretable (see Supplement C).

\begin{figure}[t]
    \centering
    \begin{subfigure}[b]{0.5\textwidth}
     \centering
        \hspace{-0.5cm}\includegraphics[scale=0.545]{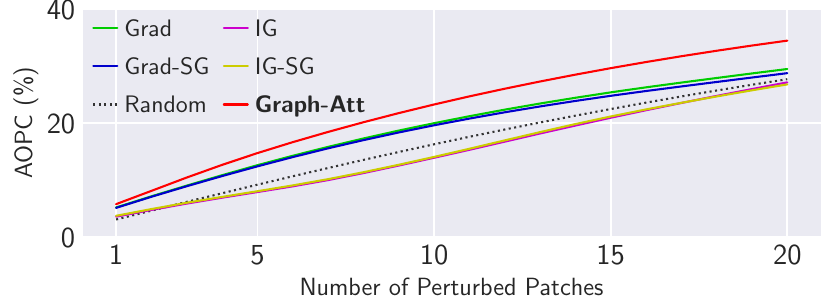}
        \caption{On F-MNIST Dataset}
    \end{subfigure}
    \begin{subfigure}[b]{0.5\textwidth}
     \centering
        \hspace{-0.5cm}\includegraphics[scale=0.545]{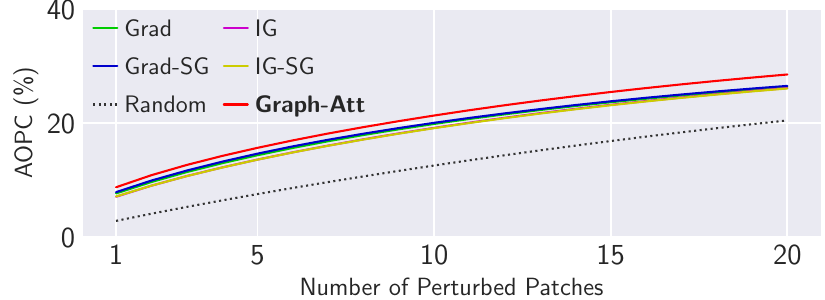}
        \caption{On CIFAR10 Dataset}
    \end{subfigure}
    \caption{Quantitative evaluation of explanations with AOPC metric: Our Graph-Att performs the best.}
    \label{fig:quant_exp}
\end{figure}

\textbf{Quantitative Evaluation} The quantitative evaluation of saliency map explanations is still an open research topic  \cite{sturmfels2020visualizing}. In this work, we quantitatively evaluate explanations with a widely used metric, i.e. Area Over the Perturbation Curve (AOPC) \cite{Samek2017EvaluatingTV} $AOPC = \frac{1}{L+1} \langle \sum^L_{k=1} f(\mathbf{X}^{(0)})- f(\mathbf{X}^{(k)})\rangle_{p(\mathbf{X})}$, where $L$ is the number of pixel deletion steps, $f(\cdot)$ is the model, $\mathbf{X}^{(K)}$ is the input image after $k$ perturbation steps. The order of perturbation steps follow the relevance order of corresponding input pixels in explanations. In each perturbation step, the target pixel is replaced by a patch ($5\times5$) with random values from $[0, 1]$. The higher the AOPC is, the more accurate the explanation are.

The AOPC scores are shown in Figure \ref{fig:quant_exp}. The difference between the baseline methods and their smoothed versions is small since our model is robust to input random perturbation noise. Our Graph-Att achieve better scores than other explanations (more results in Supplement D). On F-MNIST dataset, IG is not better than Grad, even worse than Random. The existing advanced interpretation methods are not suitable for capsule-type networks. For more methods SquaredGrad and VarGrad \cite{adebayo2018sanity}, our methods are orthogonal to them and can also be combined with them.

\begin{figure*}[t]
    \centering
    \begin{subfigure}[b]{0.24\textwidth}
        \includegraphics[width=0.85\textwidth]{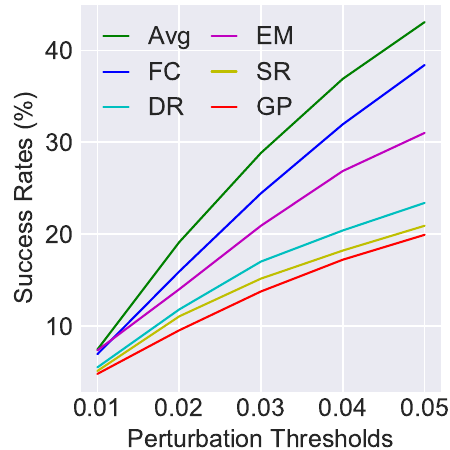}
        \caption{FGSM on SVHN}
        \label{fig:svhn_fgsm_robust}
    \end{subfigure}
    \begin{subfigure}[b]{0.24\textwidth}
        \includegraphics[width=0.85\textwidth]{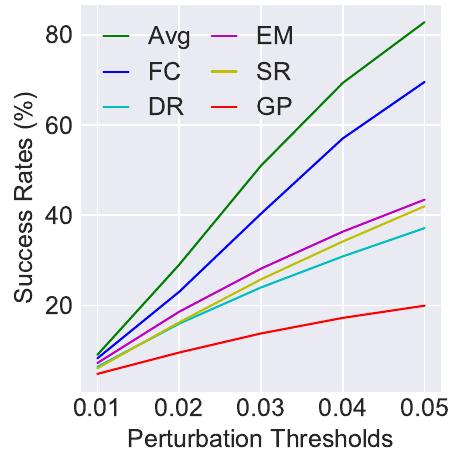}
        \caption{C\&W on SVHN}
         \label{fig:svhn_cw_robust}
    \end{subfigure} \hspace{0.2cm}
    \begin{subfigure}[b]{0.24\textwidth}
        \includegraphics[width=0.85\textwidth]{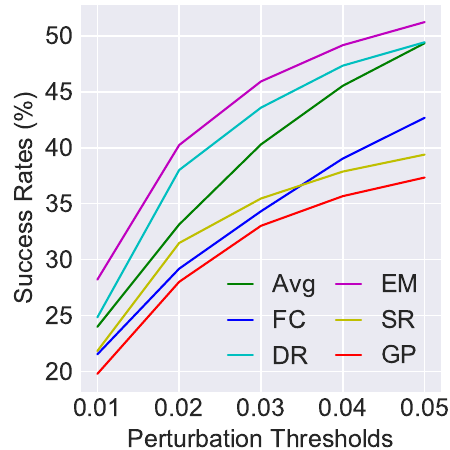}
        \caption{FGSM on CIFAR10}
         \label{fig:cifar10_fgsm_robust}
    \end{subfigure}
    \begin{subfigure}[b]{0.24\textwidth}
        \includegraphics[width=0.85\textwidth]{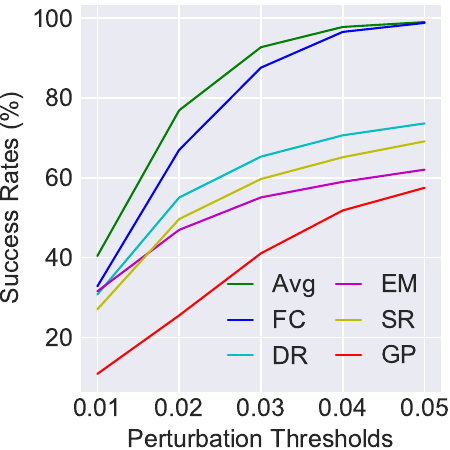}
        \caption{C\&W on CIFAR10}
         \label{fig:cifar10_cw_robust}
    \end{subfigure}
    \caption{On SVHN and CIFAR10, the attack methods attack our models (GP) with less success rate.}
    \label{fig:robust}
\end{figure*}

\textbf{Efficiency} In GraCapsNets, the single explanation created by visual attention can be obtained in half forward pass without backpropagation. Grad requires a single forward and backward pass. IG interpolates examples between a baseline and inputs, which requires M(=50) times forward and backward passes. SG variants achieve smoothy explanation by adding different noise into inputs, which require N(=10) times more forward and backward passes, i.e., N*M(=500) for IG-SG. In summary, the explanations inside our GraCapsNets is better and require less computational cost.

\subsection{Adversarial Robustness}
\label{sec:robust}
The work \cite{NIPS2019_8982} also claims that their routing mechanism is more robust to adversarial attacks. Follow their settings, we compare our model with routing algorithms in terms of the adversarial robustness.

In this experiment, we use the trained models in Section \ref{sec:sota}. FGSM \cite{Goodfellow2015ExplainingAH} (a primary attack method) and C\&W \cite{Carlini2017Robust} are applied to create adversarial examples. Their hyperparameter settings are default in Adversarial Robustness 360 Toolbox \cite{nicolae2018adversarial} (See Supplement E). The same settings are used to attack all models. Instead of choosing a single perturbation threshold, we use different thresholds, i.e., in the range $[0.01, 0.05]$ with the interval of $0.01$.

Attack success rate is used to evaluate the model robustness. Only correctly classified samples are considered in this experiment. An untargeted attack is successful when the prediction is changed, and a targeted attack is successful if the input is misclassified into the target class.

Figure \ref{fig:robust} shows the success rates of CNNs (Avg, FC), CapsNets (DR, EM, SR) and our GraCapsNets (GP) under untargeted setting. Overall, CapsNets with various routing algorithms more robust than CNNs. Especially, when the strong attack C\&W is used under a large threshold of $0.05$, all the predictions of CNNs can be misled by perturbations. The attack methods achieve less success rate on our models (GP). The experiments on the targeted setting also show similar results (See Supplement F). In our models, the attention-based graph pooling process can filter out part of noisy input features, which makes successful attacks more difficult.

\begin{figure}[h]
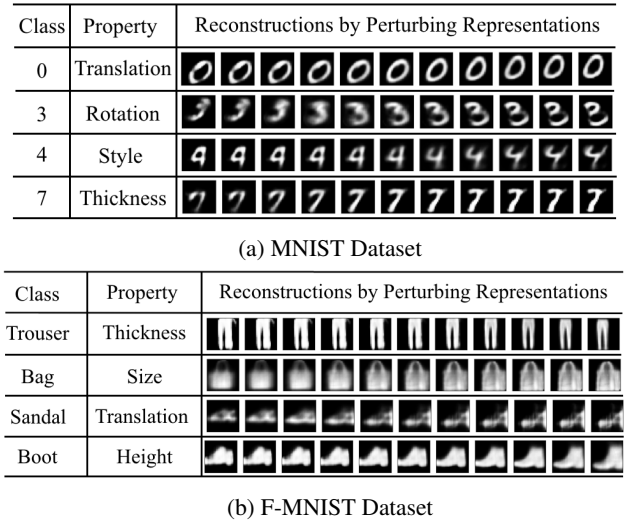

    \centering
    \begin{subfigure}[b]{0.5\textwidth}
     \centering
     \hspace{-0.5cm}
        \includegraphics[scale=0.48]{figures/mnist_rep}
        \caption{MNIST Dataset}
        \label{fig:withNoR}
    \end{subfigure}
    \begin{subfigure}[b]{0.5\textwidth}
     \centering
      \hspace{-0.5cm}
        \includegraphics[scale=0.46]{figures/fmnist_rep}
        \caption{F-MNIST Dataset}
        \label{fig:withDR}
    \end{subfigure}
    \caption{Disentangled Individual Dimensions of Representations in GraCapsNets: By perturbing one dimension of an activity vector, the variations of an input image are reconstructed.}
    \label{fig:disentanged_caps}
\end{figure}

\subsection{Disentangled Representations and Transformation Robustness}
In CapsNets, the reconstruction net reconstructs the original inputs from the disentangled activity vectors of the output capsules. When elements of the vector representation are perturbated, the reconstructed images are also changed correspondingly. We also conduct the perturbation experiments on output capsules of GraCapsNet. Similarly, we tweak one dimension of capsule representation by intervals of $0.05$ in the range $[-0.25, 0.25]$. The reconstructed images are visualized in Figure \ref{fig:disentanged_caps}. We can observe that our GraCapsNet also captures disentangled representations. For instance, the property \textit{Size} of the class \textit{Bag} in F-MNIST.

On the affine transformation benchmark task, where models are trained on the MNIST dataset and tested on the AffNIST dataset (novel affine transformed MNIST images), the CapsNets are shown to be more robust to input affine transformations than similar-sized CNNs (79\% vs. 66\%) \cite{sabour2017dynamic}. Following their setting, we also test our GraCapsNet on this benchmark, the test performance on AffNIST dataset is slightly better (80.45\%). 

\section{Conclusion}
We propose an interpretable GraCapsNet. The explanations for individual classifications of GraCapsNets can be created in an effective and efficient way. Surprisingly, without a routing mechanism, our GraCapsNets can achieve better classification performance and better adversarial robustness, and still keep other advantages of CapsNets. This work also reveals that we cannot attribute the advantages of CapsNets to the routing mechanisms, even though they are fundamental parts of CapsNets.

\bibliography{aaai21}

\end{document}